\documentclass[letterpaper]{article} 
\usepackage{aaai2026}  
\usepackage{times}  
\usepackage{helvet}  
\usepackage{courier}  
\usepackage[hyphens]{url}  
\usepackage{graphicx} 
\usepackage{natbib}  
\usepackage{caption} 
\usepackage{algorithm}
\usepackage{algorithmic}
\usepackage{multirow}
\usepackage{amssymb}
\usepackage{amsmath}
\usepackage{adjustbox}
\usepackage{newfloat}
\usepackage{listings}
\DeclareCaptionStyle{ruled}{labelfont=normalfont,labelsep=colon,strut=off} 
\floatstyle{ruled}
\newfloat{listing}{tb}{lst}{}
\floatname{listing}{Listing}
\title{Aligning the True Semantics: Constrained Decoupling and Distribution Sampling for Cross-Modal Alignment}
\author{
     Xiang Ma\textsuperscript{\rm 1},
     Lexin Fang\textsuperscript{\rm 1},
     Litian Xu\textsuperscript{\rm 2},
     Caiming Zhang\textsuperscript{\rm 1}
}
\affiliations{
    \textsuperscript{\rm 1}Shandong University, Jinan, China\\
    \textsuperscript{\rm 2}The University of Exeter, Exeter, United Kingdom\\

    xiangma@sdu.edu.cn, fanglexin@mail.sdu.edu.cn, lx268@exeter.ac.uk, czhang@sdu.edu.cn

}

\begin{document}

\maketitle

\begin{abstract}
Cross-modal alignment is a crucial task in multimodal learning aimed at achieving semantic consistency between vision and language. This requires that image-text pairs exhibit similar semantics. Traditional algorithms pursue embedding consistency to achieve semantic consistency, ignoring the non-semantic information present in the embedding. An intuitive approach is to decouple the embeddings into semantic and modality components, aligning only the semantic component. However, this introduces two main challenges: (1) There is no established standard for distinguishing semantic and modal information. (2) The modality gap can cause semantic alignment deviation or information loss. To align the true semantics, we propose a novel cross-modal alignment algorithm via \textbf{C}onstrained \textbf{D}ecoupling and \textbf{D}istribution \textbf{S}ampling (CDDS). Specifically, (1) A dual-path UNet is introduced to adaptively decouple the embeddings, applying multiple constraints to ensure effective separation. (2) A distribution sampling method is proposed to bridge the modality gap, ensuring the rationality of the alignment process. Extensive experiments on various benchmarks and model backbones demonstrate the superiority of CDDS, outperforming state-of-the-art methods by 6.6\% to 14.2\%.
\end{abstract}


\section{Introduction}
\label{sec:intro}
\begin{figure}
	\centering
	\includegraphics[width=1\linewidth]{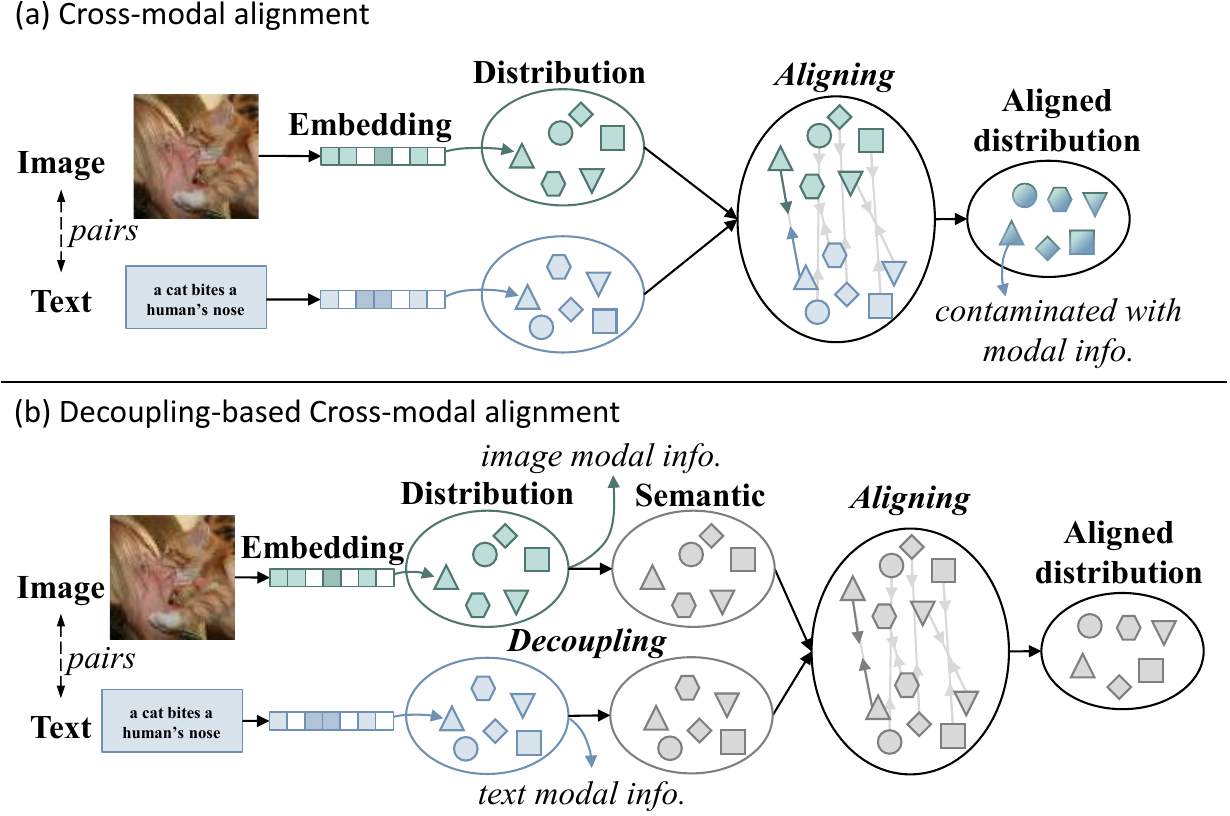}
	\caption{Comparison of cross-modal alignment and decoupling-based cross-modal alignment.}
	\label{fig1}
\end{figure}

Cross-modal alignment aims to bridge the gap between different modalities, such as vision and language. It is a fundamental technology in the field of multimodal learning, with broad applications in tasks like image-text retrieval \cite{fu2024linguistic}, image captioning \cite{guo2019image,Visual2text}, and text-to-image generation \cite{li2019visual,liao2022text}. The core of cross-modal alignment lies in ensuring semantic consistency between image-text pairs, which helps establish clear correspondences between the two modalities. Most existing state-of-the-art (SOTA) algorithms \cite{li2023your, liu2023efficient} employ contrastive learning to adjust the embeddings of image-text pairs, achieving semantic consistency through embeddings consistency, as shown in Figure \ref{fig1}(a). 

However, embeddings may contain semantically irrelevant information such as color distribution of images, syntactic structure of text, noise in training data, etc. These modality specific information cannot be matched across modalities, but the alignment process is still considering them. This introduces some erroneous implications into the alignment process, leading to deviations and even incorrect results. Consequently, embedding consistency does not guarantee semantic consistency, and aligning embeddings directly may introduce semantic bias.

An intuitive solution is to decouple embeddings into semantic and modality components, as shown in Figure \ref{fig1}(b). By aligning only the semantic components, we can avoid misleading alignment with non-semantic information. But the coupling of semantic and modality information is complex and lacks clear standards, necessitating an effective decoupling method. The main challenge is how to ensure that the semantic and non-semantic components obtained can play their corresponding roles, so as to ensure the effectiveness of decoupling. Besides, it is necessary to ensure details are not lost during the decoupling process.

Moreover, the correlation between embeddings is an important basis for achieving cross-modal alignment. The semantics represented by embeddings are mixed and implicit, but the method of constructing embeddings within the same modality is consistent. It is relatively reasonable to directly calculate embedding correlation using methods such as cosine similarity. The methods for constructing embeddings differ significantly across modalities, which is often reflected in the fact that the semantics represented by embeddings in the same column are different across modalities \cite{sidorov2014soft}. So, the contrastive learning-based methods use cosine similarity to directly interact corresponding columns of embeddings across modalities, which lacks reasonable basis. Besides, adjusting embeddings to achieve consistency can distort the original distributions of both modalities, resulting in alignment biases or information loss. Therefore, a method is needed that can achieve semantic consistency without altering the original distributions.

\begin{figure*}
	\centering
	\includegraphics[width=1\linewidth]{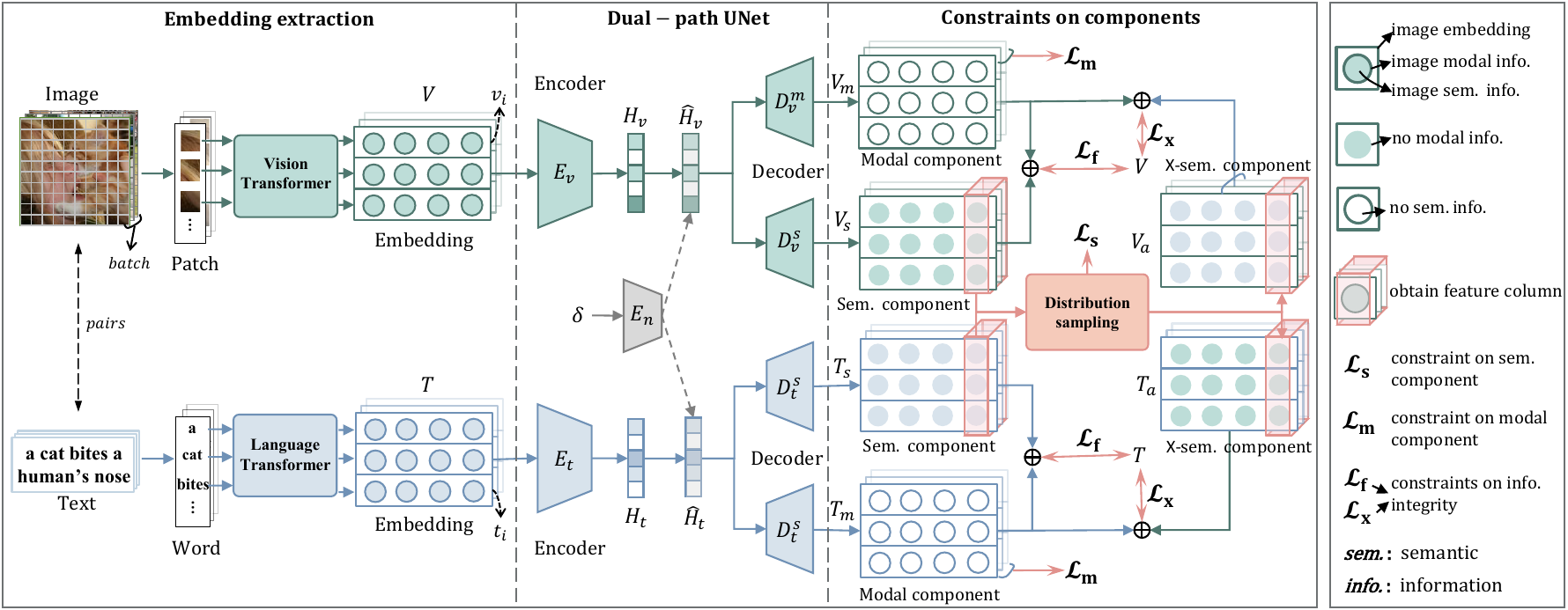}
	\caption{Overview of CDDS.}
	\label{framework}
\end{figure*}

To address the above issues, we propose a cross-modal alignment algorithm based on \textbf{C}onstrained \textbf{D}ecoupling and \textbf{D}istribution \textbf{S}ampling (\textbf{CDDS}). As shown in Figure \ref{framework}, CDDS introduces \textbf{a dual-path UNet decoupling architecture} that adaptively decouples embeddings into semantic and modality components. The key lies in applying multiple constraints on the decoupling process to ensure both the effectiveness of decoupling and the integrity of information. Specifically: (1) To ensure semantic consistency, CDDS enforces constraints on the consistency of the semantic components between image-text pairs. (2) To capture modality specific uniqueness, CDDS enforces constraints on the consistency of modality components within the same modality. (3) To maintain the information integrity during decoupling, CDDS ensures that the semantic and modality components can jointly reconstruct the original embedding. 

More importantly, we propose \textbf{a distribution sampling method} to ensure the rationality of the semantic consistency and to avoid alignment biases. This method first identifies the semantic correspondence of the embeddings for image-text pairs, to avoid erroneous guidance from unrelated semantic during the interaction. By sampling based on the distribution of related semantic in the other modality, we constructs cross-modal semantic component (x-semantic component). The x-semantic component describes the semantic information of the current modality in the description form of the other modality, effectively bridging the modality gap. By ensuring consistency between the semantic component and x-semantic component, CSSD indirectly achieves cross-modal alignment without distorting the original distributions. Our contributions are summarized as follows:
\begin{itemize}
	\item A dual-path UNet decoupling architecture is introduced to adaptively separate embeddings into semantic and modality components. Cross-modal alignment is achieved by aligning only the semantic component to ensure rationality.
	\item Multiple constraints are applied to ensure the effectiveness and the information integrity of decoupling.
	\item A distribution sampling method is proposed to indirectly achieve effective and reasonable semantic alignment.
\end{itemize}
\section{Related work}
According to the implementation, Cross-modal alignment works can be broadly categorized into coarse-grained and fine-grained methods.

\textbf{Coarse-grained methods} embed images and texts independently into a shared space, leveraging contrastive learning to align their embeddings \cite{frome2013devise, li2019visual, wang2016learning}. Previous studies within this paradigm have frequently enhanced the joint embedding space by introducing new losses \cite{faghrivse, chun2021probabilistic}, designing new architectures for each modality backbones \cite{wen2020learning, wu2019learning}, or learning better pooling methods \cite{chen2021learning, li2022multi}. VSE++ \cite{faghrivse} proposes a triplet loss with hard negative mining, which adopted by many following works. GPO \cite{chen2021learning} designs a new pooling operator that can learn from data. DIAS \cite{ma2024bridging} introduced the spatial relationships between instances, to enhance the robustness of the alignment process. 

\textbf{Fine-grained methods} perform cross-modal interaction between image patches and text words and then obtain a cumulative similarity score \cite{chen2020imram, diao2021similarity, lee2018stacked}. Previous works within this paradigm emphasize the semantic alignment between local features. For example, SCAN \cite{lee2018stacked} is the first representative work that introduces cross-attention between the two modalities to find their alignments. CAAN \cite{zhang2020context} refines this concept by introducing an additional intra-modal interaction step following the cross-modal interaction. NAAF \cite{zhang2022negative} promotes dissimilarity between mismatched patches and words to boost similarity matching. CHAN \cite{pan2023fine} proposes a method that can ignore redundant misalignments.

However, these methods still assumes that the embeddings from different modalities interact with the same semantics during correlation computation. In contrast, we focus on decoupling and aligning the related semantic of embeddings to enhance the rationality of alignment.
\section{Methodology}
Considering effectiveness and interpretability, CDDS adopts the fine-grained method. In this section, we introduce the constrained decoupling as shown in Figure \ref{framework}. 
\subsection{Constrained decoupling} \label{decoup}
\subsubsection{Embedding extraction} 
Formally, given an image $V$, we divide it into $n_v$ non-overlapping patches, and employ the vision transformer (ViT) \cite{alexey2020image, xu2022evo, vaswani2017attention} to extract visual embeddings of patches $V=\{v_i|i\in [1,n_v], v_i\in\mathbb{R}^{d}\}$. $v_i$ is the embedding of $i$-th patch. $d$ is the feature size. Similarly, given a text $T$, we employ BERT \cite{devlin2018bert} to extract textual embeddings of words $T=\{t_j|j\in [1,n_t], t_j\in\mathbb{R}^{d}\}$. $t_j$ is the embedding of $j$-th words. $n_t$ is the number of words.
\subsubsection{Dual-path UNet architecture}
To avoid semantic bias caused by the pursuit of embedding consistency, we decouple the image/text embeddings into semantic and modal components. Cross-modal alignment is achieved by ensuring the consistency of semantic component. 

The challenge arises from the complex coupling relationship between semantic and modal information, and there lacks the intuitive standards to decouple them. So we propose a dual-path U-Net architecture \cite{zhang2023skilful} for adaptive decoupling. This architecture has a shared encoder for mapping embeddings to high-dimensional representations, a semantic decoder for learning semantic component, a modal decoder for learning modal component. To ensure the robustness of the decoding process, we introduce Gaussian noise into the representations, expanding them from a deterministic values into a distributions. The decoder then analyzes these perturbed representations to obtain more comprehensive and robust embeddings. 

Taking image embedding $V=\{v_i|i\in[1,n_v],v_i\in\mathbb{R}^{d}\}$ as an example, we use ViT as the encoder to map $V$ to high-dimensional space. The encoder captures the contextual information of the input embeddings. The multi-layer nonlinear transformations allow the representations to become more abstract in the high-dimensional space, making the subsequent decoupling process more flexible, formally:
\begin{equation}
	H_v = E_{v}(V),
\end{equation}
where $H_v=\{h_i^{v}|i\in [1,n_v], h_i\in\mathbb{R}^{d}\}$ is the representation of $V$, and $h_i^{v}$ is the representation of $v_i$. $E_{v}$ is $n$ layers ViT. Then, $z$ groups of Gaussian noise $\Delta=\{\delta_i|i\in[1,z],\delta_i\in\mathbb{R}^{n_v\times d}\}$ are randomly generated and introduced to $H_v$:
\begin{equation}
	\hat{H}_i^{v} = H_v + E_{n}(\delta_i),
\end{equation}
where $\hat{H}_i^{v}$ is the perturbed representation by $\delta_i$. $E_{n}$ is a ViT layer to impose structural characteristics on the noise information, aligning it with the features constructed by the Transformer architecture. This helps to mitigate instability in the training process caused by the perturbations. We can obtain a group of perturbed representations $\hat{H}_v=\{\hat{H}_i^{v}|i\in[1,z],\hat{H}_i^{v}\in\mathbb{R}^{n_v\times d}\}$ with the same processing. 

The decoder independently analyzes each perturbed representation in $\hat{H}_v$ to extract information of semantics or modality. Each decoder layer also consider the output of the corresponding encoder layer to preserve and leverage features of the same level of abstraction. This is accomplished through skip connections between the corresponding encoder and decoder layers. We use a semantic decoder and a modality decoder to learn semantic and modal components respectively. Both decoders operate in a similar manner. Here, we take the modal decoder as an example:
\begin{equation}
	V^m_{i} = D^{m}_v(\hat{H}_i^{v}),
\end{equation}
where $V^m_{i}\in \mathbb{R}^{n_v\times d}$ represents the modal component constructed by the modal decoder through the analysis of the perturbed representation $\hat{H}_i^{v}$. $D^{m}_v$ has $n$ layers, each containing a ViT layer followed by a linear layer. The linear layer is used to map the feature size to $d$. To ensure robustness, we take the average of $\{V^m_{i}\}_{i=1}^{z}$ to construct the final modal component:
\begin{equation}
	V_m = \frac{\sum_{i=1}^{z}(V^m_{i})}{z},
\end{equation}
where $V_m=\{v_i^{m}|i\in[1,n_v],v_i^{m}\in\mathbb{R}^{d}\}$ is the modal component of image $V$, and $v_i^{m}$ corresponds to the $i$-th patch.

Through a similar process, we can obtain the semantic component of image $V_s=\{v_i^{s}|i\in[1,n_v],v_i^{s}\in\mathbb{R}^{d}\}$, the semantic component of text $T_s=\{t_i^{s}|i\in[1,n_t],t_i^{s}\in\mathbb{R}^{d}\}$, and the modal component of text $T_m=\{t_i^{m}|i\in[1,n_t],t_i^{m}\in\mathbb{R}^{d}\}$. To ensure the effectiveness, we impose constraints on the semantic and modal components, while also constraining the relationship between these components to preserve the information integrity.

\subsection{The constraint on semantic component} \label{consem}
Existing methods typically use contrastive learning to pull the semantic component vectors closer, with the core operation being the computation of correlations between image-text pairs. Effective correlation requires two embeddings providing descriptions of the semantic under a similar form. The semantic components from different modalities often emphasize different aspects of semantic, making it challenging to align information involved in correlation calculation. 

We propose the novel alignment algorithm based on distribution sampling, which identifies and aligns the semantic components of different modalities that describe the related semantics, ensuring rationality. 
\subsubsection{Related semantics identification} 
\begin{figure}
	\centering
	\includegraphics[width=0.95\linewidth]{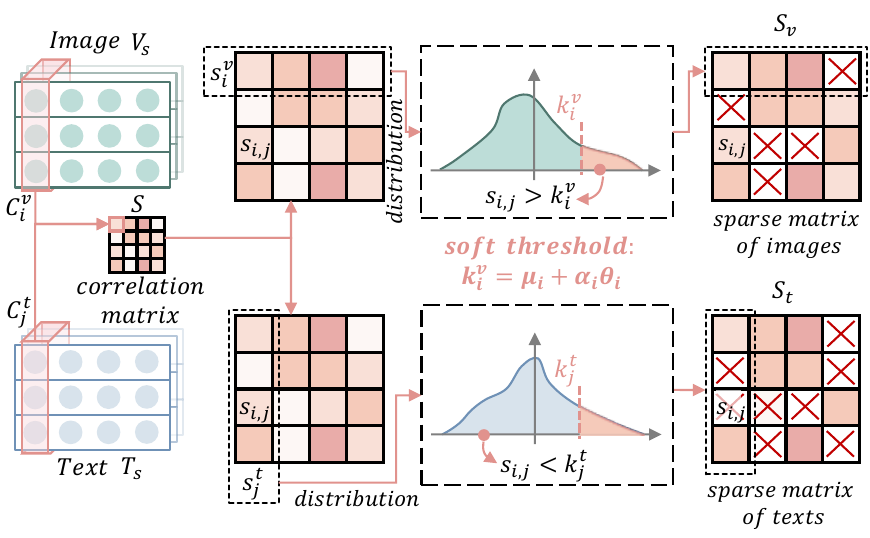}
	\caption{Related semantics identification.}
	\label{sparse}
\end{figure}

As shown in Figure \ref{sparse}, we first obtain the distribution at each feature column of all image patches and text words, and define them as $C_{v}=\{C^{v}_{i}|i\in[1,d]\}$ and $C_{t}=\{C^{t}_{j}|j\in[1,d]\}$, respectively. $C^{v}_{i}$ is the distribution estimated by values in the $i$-th column of images' semantic component, and $C^{t}_{j}$ is the distribution estimated by values in the $j$-th column of semantic component from all text. The correlation between $C^{v}_{i}$ and $C^{t}_{j}$ is obtained to determine whether they describe the related semantic:
\begin{equation}
	s_{i,j} = exp^{-\sigma_{kl}(C^{v}_{i},C^{t}_{j})},
	\label{corrfunc}
\end{equation}
A higher value of $s_{i,j}$ indicates stronger correlation between $C^{v}_{i}$ and $C^{t}_{j}$, suggesting that their descriptions are more likely to represent the related semantic. $\sigma_{kl}(\cdot)$ is the Kullback-Leibler Divergence. The correlation matrix $S=\{s_{i,j}|i,j\in[1,d]\}$ can be obtained via Eq.\ref{corrfunc}.

Next, we need to determine which distributions describe related semantics. Formally, the algorithm sparsifies each row and column of matrix $S$, retaining the larger values. Since the semantic descriptions of different modalities may not correspond one-to-one, we retain the top-$k$ distributions with the strongest correlations. The challenge lies in the significant variation in the correlation values of different distributions, making the use of a fixed $k$ as a hard-threshold inappropriate. Therefore, we propose an adaptive soft-threshold sparsification algorithm.

As shown in Figure \ref{sparse}, we define $i$-th row of $S$ as $s_{i}^{v}=\{s_{i,j}|j\in[1,d]\}$, which indicates the correlation between $C^{v}_{i}$ and all distributions of text. Define $j$-th column of $S$ as $s_{j}^{t}=\{s_{i,j}|i\in[1,d]\}$, which indicates the correlation between $C^{t}_{j}$ and all distributions of image. The strong correlation between distributions is asymmetric. If $C^{t}_{j}$ is the strong correlation distribution of $C^{v}_{i}$, $C^{v}_{i}$ may not necessarily be the strong correlation distribution of $C^{t}_{j}$. Thus, $S$ requires separate sparsification for its rows and columns. Taking row sparsification as an example, we represent the conditional probability of each column to explicitly quantification:
\begin{equation}
	p(s_{i}^{v}|s_{j}^{t}) = sigmoid(s_{i,j}),\ j\in[1,d],
\end{equation}
A higher value of $p(s_{i}^{v}|s_{j}^{t})\in[0,1]$ means the stronger correlation probability of $C^{v}_{i}$ on $C^{t}_{j}$. Based on the latent semantics of $S$, we expect the model to discover the strong correlation between distributions from different modality as concise as possible. Specifically, leveraging the statistical properties of conditional probability $\{p(s_{i}^{v}|s_{j}^{t})\}_{j=1}^{d}$, we can enable the model to learn a soft-threshold for identifying strong correlation distributions:
\begin{equation}
	k_{i}^{v} = \mu_{i}+\alpha_{i} \cdot \theta_{i},
\end{equation}
where $k_{i}^{v}$ is the soft-threshold of $s_{i}^{v}$. $\mu_{i}$ and $\theta_{i}$ are the mean and standard deviation of the sampling probability values from $\{p(s_{i}^{v}|s_{j}^{t})\}_{j=1}^{d}$, respectively. $\alpha_{i}$ is a learnable parameter to control the sparsity level.

We combine all soft-thresholds as $K_{v}=\{k_{i}^{v}|i\in [1,d]\}$, and obtain the sparse matrix:
\begin{equation}
	S_{v} = B_vS,
\end{equation}
where $S_{v}=\{s_{i,j}^{v}|i,j\in [1,d]\}$. $B_v=\{b_{i,j}^{v}|i,j\in[1,d]\}$ is a binary mask matrix to filter, and:
\begin{equation}
	s_{i,j}^{v}= \frac{b_{i,j}^{v}s_{i,j}}{\sum_{k=1}^{d}b_{i,k}^{v}s_{i,k}}, \ \ 	b_{i,j}^{v} = \left\{
	\begin{array}{lll}
		1 &s_{i,j}>k_{i}^{v}\\
		0 &otherwise
	\end{array} \right.,
\end{equation}
If $b_{i,j}=1$, then $C^{t}_{j}$ is one of the strong correlation distribution of $C^{v}_{i}$, and the correlation degree is $s_{i,j}^{v}$. That means $C^{t}_{j}$ has the related semantic as $C^{v}_{i}$.
\subsubsection{Distribution sampling} 
After capturing the strong correlations, existing methods often employ contrastive learning to adjust the distributions in pursuit of semantic consistency. This operation can distort the distributions, leading to information bias and loss. In contrast, we propose a distribution sampling method to align semantics in an indirect manner.

\begin{figure}
	\centering
	\includegraphics[width=0.95\linewidth]{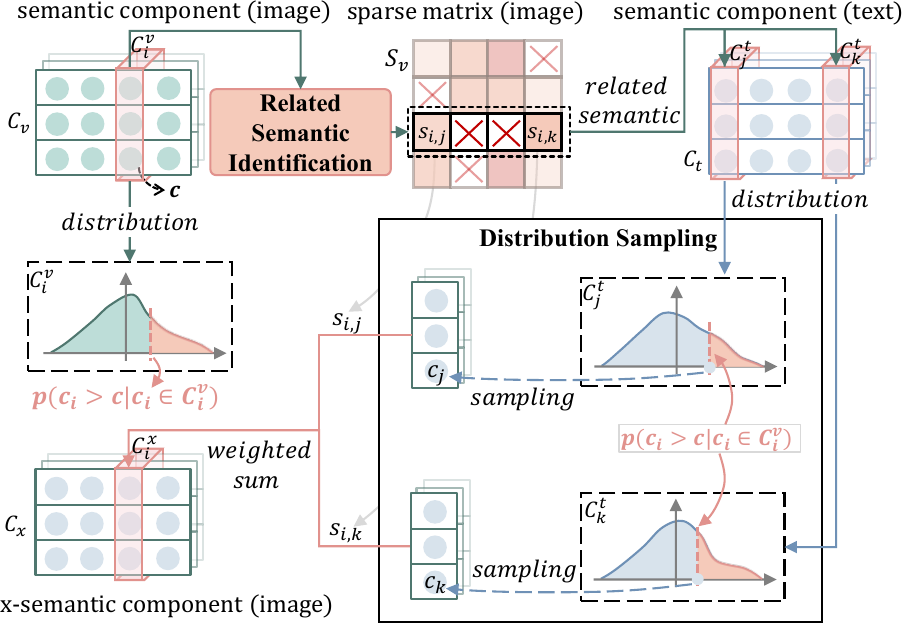}
	\caption{Distribution sampling method.}
	\label{xsem}
\end{figure}

For distribution $C^{v}_{i}$ from image modality, the method first constructs a new distribution by aggregating strong correlation distribution of text modality:
\begin{equation}
	C^{x}_{i}= \sum_{j=1}^{d} s_{i,j}^{v}\mathcal{M}_{sa}(C^{t}_j|C^{v}_{i})=\sum_{j=1}^{d}s_{i,j}^{v}\hat{C}^{x}_{j},
\end{equation}
where $C^{x}_{i}$ is the cross-modal semantic (x-semantic) corresponding to $C^{v}_{i}$. $\mathcal{M}_{sa}(C^{t}_j|C^{v}_{i})$ is the sampling operation, which constructs distribution $\hat{C}^{x}_{j}$ by sampling from the corresponding positions in $C^{t}_{j}$ based on the positions of samples in $C^{v}_{i}$, as shown in Figure \ref{xsem}. Specifically, we first calculate $p(c_i>c|c_i\in C^{v}_{i})$ for each sample $c\in C^{v}_{i}$, which is the probability of the sampling values in $C^{v}_{i}$ greater than $c$. Then we can find the value $c_j\in C^{t}_j$, which meets $p(c_i>c_j|c_i,c_j\in C^{t}_j)=p(c_i>c|c_i,c\in C^{v}_{i})$. $c_j$ is the sampling value corresponding to $c$. We can obtain $\hat{C}^{x}_{j}$ by aggregating sampling values for all samples of $C^{v}_{i}$, and weighted sum them based on $S^{v}$ the similarity between distributions to construct $C^{x}_{i}$.

This process effectively describes the semantic of $C^{v}_{i}$ using the description form of the other modality, thereby bridging the modality gap. We define $C_x=\{C^{x}_{i}|i\in[1,d]\}$ as the distribution set corresponding to $C_v$. The cross-modal semantic component (x-semantic component) of images, denoted as $V_{x}=\{v^{x}_i|i\in[1,n_v],v^{x}_i\in\mathbb{R}^d\}$, can be derived from distribution $C_x$. This process is an inverse process of distribution construction.

By performing similar operations, we can obtain the x-semantic component for text $T_{x}$. Finally, we indirectly enhence the consistency of semantic components from different modalities by ensuring the consistency between $V_{x}$ and $V_{s}$, as well as $T_{x}$ and $T_{s}$, which can be expressed as:
\begin{equation}
	\begin{aligned}
		\mathcal{L}_{s}=&-log\sum_{i=1}^{n_v}\frac{exp^{(-\sigma_{cos}(v_i^{x},v_i^{s+}))}}{\sum_{j=1,j\neq i}^{n_v}exp^{(-\sigma_{cos}(v_i^{x},v_j^{s}))}}\\
		&-log\sum_{i=1}^{n_t}\frac{exp^{(-\sigma_{cos}(t_i^{x},t_i^{s+}))}}{\sum_{j=1,j\neq i}^{n_t}exp^{(-\sigma_{cos}(t_i^{x},t_j^{s}))}},
		\label{lossse}
	\end{aligned}
\end{equation}
where $\mathcal{L}_{s}$ is the regularizer for semantic consistency. Both terms are implemented based on the concept of contrastive learning \cite{hu2024comprehensive}, which involves pulling closer the embeddings of patch-word pairs while pushing apart those of non-paired ones. The $v_i^{x}\in V_x$ is obtained by non-parametric sampling from $T_s$, so the first term can indirectly achieve semantic alignment between $V_s$ and $T_s$ by aligning $V_s$ and $V_x$. Similarly, the second term can achieve semantic alignment between $V_s$ and $T_s$ by aligning $T_s$ and $T_x$. 
\subsection{The constraint on modal component} \label{conmod}
The modal component should remain consistent within the same modality to ensure that it accurately represents modality specific uniqueness. This requires constraining the distribution consistency of modal components for all patches or words, which can be achieve by:
\begin{equation}
	\mathcal{L}_{m} = \sum_{i,j=1}^{n_v}exp^{-\sigma_{kl}(v_i^{m},v_j^{m})}+\\
	\sum_{i,j=1}^{n_t}exp^{-\sigma_{kl}(t_i^{m},t_j^{m})},
\end{equation}
where $\mathcal{L}_{m}$ is the regularizer for modality consistency. The first and second terms control the consistency between the modal components of patches and words, respectively. $\sigma_{kl}$ is the Kullback-Leibler Divergence to describe the correlation between the distribution of patches or words. 
\subsection{The constraint on information integrity} \label{coninf}
To ensure the integrity of decoupled information, the modal component and semantic component should be capable of reconstructing the original embedding:
\begin{equation}
	\mathcal{L}_{f} = \sum_{i=1}^{n_v}||w_mv_i^{m} + w_sv_i^{s} - v_i||_2^2 + \sum_{i=1}^{n_t}||w_mt_i^{m} + w_st_i^{s} - v_i||_2^2,
\end{equation}
where $\mathcal{L}_{f}$ is the regularizer for information integrity, and the first and second terms apply to patches and words. 
\begin{table*}[h]
	\centering
	\resizebox{\textwidth}{!}{\setlength{\tabcolsep}{1pt}{
			\begin{tabular}{l|ccccccc|ccccccc|ccccccc}
				\hline
				\multirow{3}{*}{Methods}&\multicolumn{7}{c|}{Flickr30K}&\multicolumn{7}{c|}{MS-COCO 1K}&\multicolumn{7}{c}{MS-COCO 5K}\\
				&\multicolumn{3}{c}{Image-to-Text}&\multicolumn{3}{c}{Text-to-Image}&rSum&\multicolumn{3}{c}{Image-to-Text}&\multicolumn{3}{c}{Text-to-Image}&rSum&\multicolumn{3}{c}{Image-to-Text}&\multicolumn{3}{c}{Text-to-Image}&rSum\\
				&R@1&R@5&R@10&R@1&R@5&R@10&&R@1&R@5&R@10&R@1&R@5&R@10&&R@1&R@5&R@10&R@1&R@5&R@10&\\
				\hline
				\multicolumn{22}{l}{\textbf{\textit{ViT-224}, 14$\times$14 patches}}\\
				VSE++&71.8&92.8&96.5&59.4&84.7&90.9&496.1&75.0&94.6&98.0&62.7&89.4&94.9&514.6&52.4&80.3&88.8&40.6&70.4&81.1&413.4\\
				SCAN&69.5&90.9&95.6&56.4&83.1&90.0&485.6&76.0&95.4&98.1&64.5&90.8&95.8&520.6&53.9&81.8&90.0&42.9&72.3&82.5&423.5\\
				SGR&69.7&90.8&95.2&59.1&84.1&89.9&488.7&77.2&95.0&98.0&65.1&90.7&95.8&521.8&54.9&82.8&90.5&42.8&72.2&82.5&425.8\\
				CHAN&69.2&91.8&95.0&58.4&84.9&90.6&489.9&77.1&95.1&98.1&65.0&91.0&96.0&522.2&56.3&83.2&90.1&43.0&72.6&82.8&428.0\\
				LAPS&74.0&93.4&97.4&62.5&87.3&92.7&507.3&78.7&95.5&98.3&66.2&91.3&96.2&526.3&57.5&84.0&90.8&44.5&74.0&83.6&434.4\\
				\textbf{CDDS}&\textbf{74.8}&\textbf{93.6}&\textbf{97.8}&\textbf{63.1}&\textbf{88.2}&\textbf{93.1}&\textbf{510.6}&\textbf{79.0}&\textbf{95.9}&\textbf{98.7}&\textbf{66.5}&\textbf{91.8}&\textbf{96.8}&\textbf{528.7}&\textbf{57.9}&\textbf{84.6}&\textbf{91.4}&\textbf{45.0}&\textbf{74.8}&\textbf{84.1}&\textbf{437.8}\\
				\hline
				\multicolumn{22}{l}{\textbf{\textit{ViT-384}, 24$\times$24 patches}}\\
				VSE++&77.1&95.7&97.5&65.8&90.2&94.3&520.5&77.0&95.7&98.4&64.6&91.1&96.2&523.0&54.9&82.8&90.4&42.4&72.4&82.8&425.8\\
				SCAN&75.4&94.4&96.9&63.6&88.6&93.5&512.5&76.1&95.5&98.5&65.1&91.6&96.3&523.1&53.3&81.8&90.0&42.6&72.6&82.9&423.1\\
				SGR&76.9&94.9&\textbf{98.1}&64.2&88.4&93.3&515.8&75.8&95.7&98.6&65.6&92.0&96.5&524.2&53.3&81.0&89.6&42.9&73.1&83.7&423.6\\
				CHAN&75.4&94.5&97.6&63.2&88.6&93.1&512.4&78.1&95.8&98.6&66.1&92.1&96.6&527.3&55.6&83.8&91.2&43.4&73.6&83.5&431.1\\
				LAPS&79.0&96.0&\textbf{98.1}&67.3&90.5&94.5&525.4&78.7&96.3&\textbf{98.9}&68.0&92.4&96.8&531.0&57.4&84.9&92.5&46.4&75.8&85.2&442.2\\
				\textbf{CDDS}&\textbf{79.5}&\textbf{96.3}&\textbf{98.1}&\textbf{67.5}&\textbf{90.8}&\textbf{94.6}&\textbf{526.8}&\textbf{79.1}&\textbf{96.8}&\textbf{98.9}&\textbf{68.3}&\textbf{92.6}&\textbf{96.9}&\textbf{532.6}&\textbf{57.8}&\textbf{85.2}&\textbf{92.8}&\textbf{46.9}&\textbf{76.2}&\textbf{85.5}&\textbf{444.4}\\
				\hline
				\multicolumn{22}{l}{\textbf{\textit{Swin-224}, 7$\times$7 patches}}\\
				VSE++&82.5&96.5&98.9&70.0&91.4&95.1&534.4&83.3&97.5&99.3&71.0&93.0&96.7&540.9&64.0&88.2&94.2&49.9&78.0&86.6&460.9\\
				SCAN&79.0&95.9&98.2&67.7&90.6&94.9&526.3&80.9&97.0&99.1&69.7&93.1&97.1&536.9&60.7&86.6&93.2&48.1&77.1&86.1&451.8\\
				SGR&80.4&97.0&98.7&66.9&90.2&94.5&527.6&81.2&97.1&99.1&69.9&93.2&97.2&537.7&61.0&86.7&93.2&48.6&77.2&86.3&453.1\\
				CHAN&81.4&97.0&98.6&68.5&90.6&94.5&530.6&81.6&97.2&99.3&70.6&\textbf{93.7}&97.6&539.8&64.1&87.9&93.5&49.1&77.3&86.1&458.0\\
				LAPS&82.4&97.4&99.5&70.0&91.7&95.4&536.3&84.0&97.6&99.3&\textbf{72.1}&\textbf{93.7}&97.3&544.1&64.5&89.2&94.4&51.6&78.9&87.2&465.8\\
				\textbf{CDDS}&\textbf{83.0}&\textbf{98.1}&\textbf{99.7}&\textbf{70.3}&\textbf{92.1}&\textbf{95.8}&\textbf{539.0}&\textbf{84.6}&\textbf{97.9}&\textbf{99.4}&\textbf{72.1}&93.5&\textbf{97.6}&\textbf{545.1}&\textbf{64.6}&\textbf{89.8}&\textbf{94.8}&\textbf{51.9}&\textbf{79.0}&\textbf{87.6}&\textbf{467.7}\\
				\hline
				\multicolumn{22}{l}{\textbf{\textit{Swin-384}, 12$\times$12 patches}}\\
				VSE++&83.3&97.5&99.2&71.1&93.2&96.2&540.6&82.9&97.7&99.4&71.3&93.5&97.3&542.1&63.0&88.5&94.3&50.1&78.9&87.4&462.2\\
				SCAN&81.9&96.9&98.9&70.0&92.7&95.8&536.1&81.6&96.8&99.1&69.1&92.7&96.7&536.1&61.1&87.3&93.3&47.8&76.9&85.9&452.4\\
				SGR&80.7&96.8&99.0&69.9&91.7&95.3&533.4&81.9&96.7&99.1&69.3&92.8&96.7&536.6&62.8&87.0&92.9&48.1&77.0&86.0&453.8\\
				CHAN&81.2&96.7&98.8&70.3&92.2&95.9&535.0&83.1&97.3&99.2&70.4&93.1&97.1&540.2&63.4&88.4&94.1&49.2&77.9&86.6&459.5\\
				LAPS&85.1&97.7&99.2&74.0&93.0&96.3&545.3&84.1&97.4&99.2&72.1&93.9&97.4&544.1&67.1&88.6&94.3&53.0&79.5&87.6&470.1\\
				\textbf{CDDS}&\textbf{86.8}&\textbf{98.3}&\textbf{99.6}&\textbf{76.3}&\textbf{94.3}&\textbf{97.2}&\textbf{552.5}&\textbf{84.9}&\textbf{98.0}&\textbf{99.6}&\textbf{73.5}&\textbf{94.6}&\textbf{98.0}&\textbf{548.6}&\textbf{67.9}&\textbf{89.6}&\textbf{94.9}&\textbf{51.1}&\textbf{80.2}&\textbf{88.4}&\textbf{472.1}\\
				\hline
	\end{tabular}}}
	\caption{Comparisons of performances with state-of-the-art methods. We list the backbones, image resolution, and the number of patches (e.g., `ViT-224, $14\times14$ patches' means the base-version of ViT with $224\times224$ image resolution input, getting $14\times14$ patches for one image, and the base-version of BERT for text words). The bests are in \textbf{bold}.}
	\label{sota}
\end{table*}
\begin{table}[h]
	\centering
	\resizebox{0.48\textwidth}{!}{\setlength{\tabcolsep}{3.2pt}{\begin{tabular}{l|cccc|cccc}
				\hline
				\multirow{3}{*}{Methods}& \multicolumn{4}{c|}{Flickr30K}  & \multicolumn{4}{c}{MS-COCO 5K}\\
				&\multicolumn{2}{c}{I-to-T}&\multicolumn{2}{c|}{T-to-I}&\multicolumn{2}{c}{I-to-T}&\multicolumn{2}{c}{T-to-I}\\
				&R@1&R@5&R@1&R@5&R@1&R@5&R@1&R@5\\
				\hline
				VILT&83.5&96.7&64.4&88.7&61.5&86.3&42.7&72.9\\
				SOHO&86.5&98.1&72.5&92.7&66.4&88.2&50.6&78.0\\
				ALBEF&95.9&99.8&85.6&97.5&77.6&94.3&60.7&84.3\\
				BLIP&96.6&99.8&87.2&97.5&80.6&95.2&63.1&85.3\\
				\hline
				\multicolumn{9}{l}{\textbf{\textit{CLIP(ViT-224 + BERT)}, 14$\times$14 patches}}\\
				CLIP$^{\#}$&81.4&96.2&61.1&85.4&52.3&76.2&33.3&58.2\\
				VSE++&92.2&99.1&80.5&95.6&66.8&88.2&53.6&79.7\\
				SCAN&88.2&98.1&75.3&93.1&65.4&88.0&50.7&77.6\\
				LAPS&92.9&99.3&80.6&95.5&69.8&90.4&54.3&80.0\\
				\textbf{CDDS} &\textbf{93.5}&\textbf{99.6}&\textbf{81.2}&\textbf{96.4}&\textbf{70.4}&\textbf{91.0}&\textbf{54.8}&\textbf{80.9}\\
				\hline
				\multicolumn{9}{l}{\textbf{\textit{CLIP(ViT-224-Large + BERT-Large)}, 16$\times$16 patches}}\\
				CLIP$^{\#}$&85.0&97.7&64.3&87.0&55.9&79.1&35.9&60.9\\
				VSE++&94.0&99.5&83.4&96.4&68.5&89.4&56.7&81.9\\
				SCAN&90.0&98.5&81.0&95.9&68.0&90.4&53.2&80.7\\
				LAPS&94.6&\textbf{99.9}&84.9&97.3&72.9&91.7&57.1&81.3\\
				\textbf{CDDS}&\textbf{95.2}&\textbf{99.9}&\textbf{85.3}&\textbf{98.0}&\textbf{73.3}&\textbf{92.2}&\textbf{57.6}&\textbf{81.6}\\
				\hline
	\end{tabular}}}
	\caption{Comparisons with Vision-Language Pre-training (VLP) Models. `$\#$' denotes the zero-shot learning.}
	\label{sota2}
\end{table}
To avoid information loss caused by semantic alignment, we can also constrain the consistency between the combination of modal and x-semantic components with the original embedding:
\begin{equation}
	\mathcal{L}_{x} = \sum_{i=1}^{n_v}||w_mv_i^{m} + w_xv_i^{x} - v_i||_2^2 + \sum_{i=1}^{n_t}||w_mt_i^{m} + w_xt_i^{x} - v_i||_2^2,
\end{equation}
where $w_m$, $w_x$ and $w_s$ are learnable parameters.
\subsection{Objective function} 
We combine the regularizers to obtain the loss function of CDDS:
\begin{equation}
	\mathcal{L} = \alpha_s\mathcal{L}_{s} + \alpha_m\mathcal{L}_{m} + \alpha_f\mathcal{L}_{f} + (1-\alpha_f)\mathcal{L}_{x},
	\label{fff}
\end{equation}
where $\alpha_s$, $\alpha_m$, $\alpha_f$ and $\alpha_i$ are hyper-parameters to control the effectiveness degree of regularizers. There is an interdependence between $\mathcal{L}_{f}$ and $\mathcal{L}_{x}$, and $\alpha_f$ can adjust whether to use semantic component or x-semantic component to ensure information integrity. At the inference stage, it is sufficient to decouple the input and calculate correlations between semantic components to determine the matching relationship without any other operations.
\section{Experiments}
\subsection{Datasets and evalution metrics}
Following the previous works \cite{fu2024linguistic, ma2024bridging}, we evaluate CDDS on the typical Flickr30K \cite{young2014image} and MS-COCO \cite{lin2014microsoft} datasets. Flickr30k contains 29,000 images for training, 1,000 images for validation, and 1,000 images for testing. MSCOCO contains 82,738 images for training, 5,000 images for validation, and 5,000 images for testing. Each image is associated with 5 texts. The results on MS-COCO are reported on averaging over 5-folds of 1,000 test images and on the full 5,000 test images. The evaluation metrics are the Recall at K (R@K) and rSum. R@K means the percentage of ground truth in the retrieved top-K lists, and K=1,5,10. rSum is the sum of multiple R@K in both image-to-text and text-to-image, which reflects the overall performance.
\subsection{Implementation details}
We use the base version Vision Transformer (ViT) \cite{alexey2020image} and Swin Transformer (Swin) \cite{liu2021swin} as backbones to extract visual embeddings, and use BERT \cite{devlin2018bert} to extract textual embeddings. A patch is $16\times16$ pixels for ViT, and is $32\times32$ pixels for Swin. The image resolutions are $224\times224$ and $384\times384$. So we obtain $14\times14$ and $24\times24$ patches for ViT, and obtain $7\times7$ and $12\times12$ patches for Swin. An additional linear layer is introduced on the top of these backbones to unify feature size $d$ as 512. The whole framework is trained for 25 epochs on a NVIDIA L40 GPU. AdamW optimizer \cite{loshchilov2017decoupled, zhang2018improved} is adopted with learning rate of $2e^{-4}$. The batch size is 64. The layer number is 2 for encoders and decoders.
\begin{table}[t]
	\centering
	\resizebox{0.48\textwidth}{!}{\setlength{\tabcolsep}{1.8pt}{\begin{tabular}{l|cccc|cccc|c}
				\hline
				\multirow{3}{*}{Methods}& \multicolumn{4}{c|}{Flickr30K}  & \multicolumn{4}{c|}{MS-COCO 5K}&\\
				&\multicolumn{2}{c}{I-to-T}&\multicolumn{2}{c|}{T-to-I}&\multicolumn{2}{c}{I-to-T}&\multicolumn{2}{c|}{T-to-I}&$CR$\\
				&R@1&R@5&R@1&R@5&R@1&R@5&R@1&R@5&\\
				\hline
				\multicolumn{10}{l}{\textbf{\textit{ViT-224}, 14$\times$14 patches}}\\
				w/o $Dec.$&71.4&91.8&57.9&82.5&54.8&82.6&43.2&72.2&-4.6\%\\
				w/o $Mod.$&73.8&93.2&62.3&86.9&57.6&84.1&44.6&74.1&-0.9\%\\
				w/o $Int.$&69.8&91.5&57.9&80.4&52.1&81.2&42.0&70.8&-6.7\%\\
				w/o $Gau.$&74.2&93.5&62.9&87.8&57.8&84.3&44.6&74.5&-0.4\%\\
				w/o $Sam.$&72.1&92.0&60.4&86.9&56.9&83.5&44.1&73.8&-2.1\%\\
				\textbf{CDDS}&\textbf{74.8}&\textbf{93.6}&\textbf{63.1}&\textbf{88.2}&\textbf{57.9}&\textbf{84.6}&\textbf{45.0}&\textbf{74.8}&--\\
				\hline
				\multicolumn{10}{l}{\textbf{\textit{Swin-224}, 7$\times$7 patches}}\\
				w/o $Dec.$&82.2&97.4&70.0&91.4&64.1&88.9&51.4&78.8&-0.7\%\\
				w/o $Mod.$&80.4&93.9&65.1&90.2&61.8&86.0&48.2&77.7&-4.2\%\\
				w/o $Int.$&80.3&93.4&65.1&89.2&61.2&85.1&47.9&77.2&-4.9\%\\
				w/o $Gau.$&82.6&97.8&70.2&91.9&64.5&89.4&51.7&78.9&-0.3\%\\
				w/o $Sam.$&81.7&97.2&70.0&91.5&63.9&88.8&51.2&78.5&-1.0\%\\
				\textbf{CDDS}&\textbf{83.0}&\textbf{98.1}&\textbf{70.3}&\textbf{92.1}&\textbf{64.6}&\textbf{89.8}&\textbf{51.9}&\textbf{79.0}&--\\
				\hline
	\end{tabular}}}
	\caption{Ablation studies of CDDS. $CR$ represents the rate of change in the effect of removing the module.}
	\label{ab}
\end{table}
\subsection{Comparisons with state-of-the-art methods}
To show the performance superiority of CDDS, we compare it on the two datasets with five state-of-the-art (SOTA) methods: VSE++\cite{faghrivse}, SCAN\cite{lee2018stacked}, SGR\cite{diao2021similarity}, CHAN\cite{pan2023fine}, and LAPS\cite{fu2024linguistic}. The results are cited directly from the paper of LAPS\cite{fu2024linguistic}. 

As shown in Table \ref{sota}, CDDS outperformers SOTA methods with impressive margins for the R@K and rSum, and achieves superiority on different backbones. Notably, the performance improves with more complex transformer-based backbones. This complexity is reflected in the type of backbone used and the number of patches processed.

Additionally, we extend the our architecture to the classic vision-language pre-training (VLP) model CLIP \cite{radford2021learning} shown in Table \ref{sota2}. By comparing it with the current SOTA VLP models (VILT\cite{chen2020uniter}, SOHO\cite{kim2021vilt}, ALBEF \cite{huang2021seeing}, BLIP\cite{li2022blip}), we find that existing fine-grained methods, even when using a VLP backbone, struggle to achieve satisfactory results. In contrast, CDDS offers significant improvements and demonstrating competitive performance compared to the mainstream VLP models.
\begin{table}[h]
	\centering
	\resizebox{0.47\textwidth}{!}{\setlength{\tabcolsep}{1.8pt}{\begin{tabular}{l|cccc|cccc|c}
				\hline
				\multirow{3}{*}{Methods}& \multicolumn{4}{c|}{Flickr30K}  & \multicolumn{4}{c|}{MS-COCO 5K}&\\
				&\multicolumn{2}{c}{I-to-T}&\multicolumn{2}{c|}{T-to-I}&\multicolumn{2}{c}{I-to-T}&\multicolumn{2}{c|}{T-to-I}&$CR$\\
				&R@1&R@5&R@1&R@5&R@1&R@5&R@1&R@5&\\
				\hline
				VSE++&71.8&92.8&59.4&84.7&52.4&80.3&40.6&70.4&\multirow{2}{*}{+0.6\%}\\
				\textbf{+$Sam$}&72.1&93.0&59.7&85.0&52.8&80.9&41.5&70.8&\\
				\hline
				SCAN&69.5&90.9&56.4&83.1&53.9&81.8&42.9&72.3&\multirow{2}{*}{+1.1\%}\\
				\textbf{+$Sam$}&69.8&91.2&56.4&83.8&54.9&82.9&44.1&73.9&\\
				\hline
				SGR&69.7&90.8&59.1&84.1&54.9&82.8&42.8&72.2&\multirow{2}{*}{+0.9\%}\\
				\textbf{+$Sam$}&70.4&91.1&59.5&85.0&55.5&83.6&43.2&73.0&\\
				\hline
				CHAN&69.2&91.8&58.4&84.9&56.3&83.2&43.0&72.6&\multirow{2}{*}{+0.4\%}\\
				\textbf{+$Sam$}&69.3&92.6&58.4&85.0&56.6&83.6&43.2&72.6&\\
				\hline
				LAPS&74.0&93.4&62.5&87.3&57.5&84.0&44.5&74.0&\multirow{2}{*}{+0.5\%}\\
				\textbf{+$Sam$}&74.3&93.4&62.8&87.9&57.9&84.2&44.9&74.7&\\
				\hline
	\end{tabular}}}
	\caption{The application effect of distribution sampling method to other models (trained with `\textbf{\textit{ViT-224}}').}
	\label{cross}
\end{table}
\begin{table}[h]
	\centering
	\setlength{\tabcolsep}{1.5pt}{\begin{tabular}{c|c|c|c}
			\hline
			&\multirow{2}{*}{Batch Time(s)}&\multicolumn{2}{c}{rSum}\\
			&&Flickr30K&MS-COCO 5K\\
			\hline
			Each batch&0.11&510.6&437.8\\
			Random &0.06&502.3&421.6\\
			All&0.02&483.6&411.6\\
			\hline
	\end{tabular}}
	\caption{The limitation in efficiency. `Each batch' means Eq.\ref{corrfunc} is executed at each batch. `Random' is the random selection of features. `All' means appling Eq.\ref{corrfunc} to the total dataset before training.}
	\label{lim}
\end{table}
\begin{figure}[t]
	\centering
	\includegraphics[width=\linewidth]{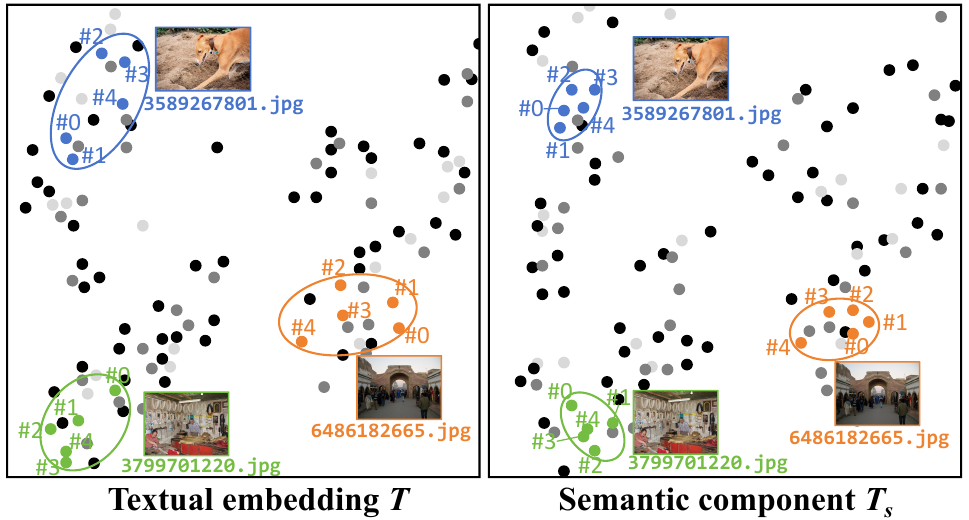}
	\caption{Visualization of textual embeddings and semantic components, showing decoupling can bring textual embeddings with similar semantics (corresponding to the same image) closer. `\#0' is the text's identifier.}
	\label{ca}
\end{figure}
\subsection{Ablation study}
To demonstrate the effectiveness of the modules in CDDS, we conduct extensive ablation studies and analyses on both datasets. The baseline w/o $Dec.$ means the removal of decoupling architecture. w/o $Mod.$ means no the modal component constraint. w/o $Int.$ means no information integrity constraint. w/o $Gau.$ means not adding Gaussian noise to high-dimensional representations. w/o $Sam.$ means replacing the distribution sampling method with contrastive learning for semantic alignment. According to the Table \ref{ab}, we can find removing any modules in CDDS results in a performance decline, indicating that the constrained decoupling architecture and distribution sampling are indeed integral to the effectiveness of cross-modal alignment. Table \ref{cross} shows that applying distribution sampling to other models also improves their performance, demonstrating its robustness.
\subsection{Visualization}
Figure \ref{ca} shows the qualitative analysis results of embedding changes before and after decoupling. It can be seen that the decoupling process removes modal information, making textual embeddings with similar semantics closer, which is beneficial for image-text retrieval.

\subsection{Limitation and discussion}
The Eq.\ref{corrfunc} needs to be performed in each batch, making the process computationally expensive ($\mathcal{O}(N^2)$). Therefore, we attempt to calculate Eq.\ref{corrfunc} across the total dataset or introduce random sampling approach to reduce the number of features that need to be computed, as shown in Tab.\ref{lim}. While these approaches significantly reduces the computational time, it has a substantial impact on the effectiveness.
\section{Conclusion}
In this paper, we propose a novel cross-modal alignment algorithm based on Constrained Decoupling and Distribution Sampling (CDDS). We introduce a dual-path UNet decoupling architecture to adaptively distinguish the semantic and modality components, applying multiple constraints to ensure the decoupling effectiveness and the information integrity. Besides, we propose a distribution sampling method to identify the semantic correspondence of different modalities, indirectly achieving cross-modal alignment without adjusting the embeddings. Extensive experiments and analyses conducted on various benchmarks and backbones demonstrate the superiority and rationality of CDDS.
\section*{Acknowledgments}
The authors appreciate the financial support by the National Natural Science Foundation of China (NSFC) Joint Fund with Zhejiang Integration of Informatization and Industrialization under Key Project (Grant Number U22A2033), the Postdoctoral Fellowship Program of CPSF under Grant Number GZC20251643, the NSFC under Grant Number 62576193.

\bibliography{main}
\end{document}